\pdfoutput=1

\documentclass[11pt]{article}

\usepackage{EMNLP2022}

\usepackage{times}
\usepackage{latexsym}
\usepackage{amsmath}
\usepackage{diagbox}
\usepackage{caption}
\usepackage{amsfonts,amssymb}
\usepackage{subfigure}
\usepackage{graphicx}
\usepackage{latexsym}
\usepackage{bm}
\usepackage{amssymb}
\usepackage{amsmath}
\usepackage{multirow}
\usepackage[ruled,lined,commentsnumbered]{algorithm2e}
\usepackage{hyperref} 
\usepackage{booktabs}
\usepackage{bbm}
\usepackage{bm}
\usepackage[T1]{fontenc}

\usepackage[utf8]{inputenc}

\usepackage{microtype}

\usepackage{inconsolata}

\title{Turning Fixed to Adaptive: Integrating Post-Evaluation 

into Simultaneous Machine Translation}

\author{
    Shoutao Guo \textsuperscript{\rm 1,2},
    Shaolei Zhang \textsuperscript{\rm 1,2},
    Yang Feng \textsuperscript{\rm 1,2}\thanks{ \ \ Corresponding author: Yang Feng.} \\
        \textsuperscript{\rm 1}{Key Laboratory of Intelligent Information Processing} \\ Institute of Computing Technology, Chinese Academy of Sciences (ICT/CAS) \\
    { \textsuperscript{\rm 2} {University of Chinese Academy of Sciences, Beijing, China}} \\
     \texttt{\{\href{mailto:guoshoutao22z@ict.ac.cn}{guoshoutao22z}, \href{mailto:zhangshaolei20z@ict.ac.cn}{zhangshaolei20z}, \href{mailto:fengyang@ict.ac.cn}{fengyang}\}@ict.ac.cn}  }

\begin{document}
\maketitle
\begin{abstract}
Simultaneous machine translation (SiMT) starts its translation before reading the whole source sentence and employs either fixed or adaptive policy to generate the target sentence. Compared to the fixed policy, the adaptive policy achieves better latency-quality tradeoffs by adopting a flexible translation policy. If the policy can evaluate rationality before taking action, the probability of incorrect actions will also decrease. However, previous methods lack evaluation of actions before taking them. In this paper, we propose a method of performing the adaptive policy via integrating post-evaluation into the fixed policy. Specifically, whenever a candidate token is generated, our model will evaluate the rationality of the next action by measuring the change in the source content. Our model will then take different actions based on the evaluation results. Experiments on three translation tasks show that our method can exceed strong baselines under all latency\footnote{Code is available at \url{https://github.com/ictnlp/PED-SiMT}}.
\end{abstract}

\section{Introduction}

Simultaneous machine translation (SiMT) \citep{gu-etal-2017-learning, DBLP:conf/acl/MaHXZLZZHLLWW19, arivazhagan-etal-2019-monotonic, DBLP:conf/iclr/MaPCPG20, zhang-feng-2021-modeling-concentrated, zhang-feng-2022-reducing} starts translation before reading the whole source sentence. It seeks to achieve good latency-quality tradeoffs and is suitable for various scenarios with different latency tolerances. Compared to full-sentence machine translation, SiMT is more challenging because it lacks partial source content in translation and needs to decide on translation policy additionally.

The translation policy in SiMT directs the model to decide when to take READ (i.e., read the next source token) or WRITE (i.e., output the generated token) action, so as to ensure that the model has appropriate source content to translate the target tokens. Because READ and WRITE actions are often decided based on available source tokens and generated target tokens, it is difficult to guarantee their accuracy. Therefore, if the SiMT model can evaluate the rationality of actions with the help of the current generated candidate token, it can reduce the probability of taking incorrect actions.

\begin{figure}[t]
    \centering
    \includegraphics[width=3in]{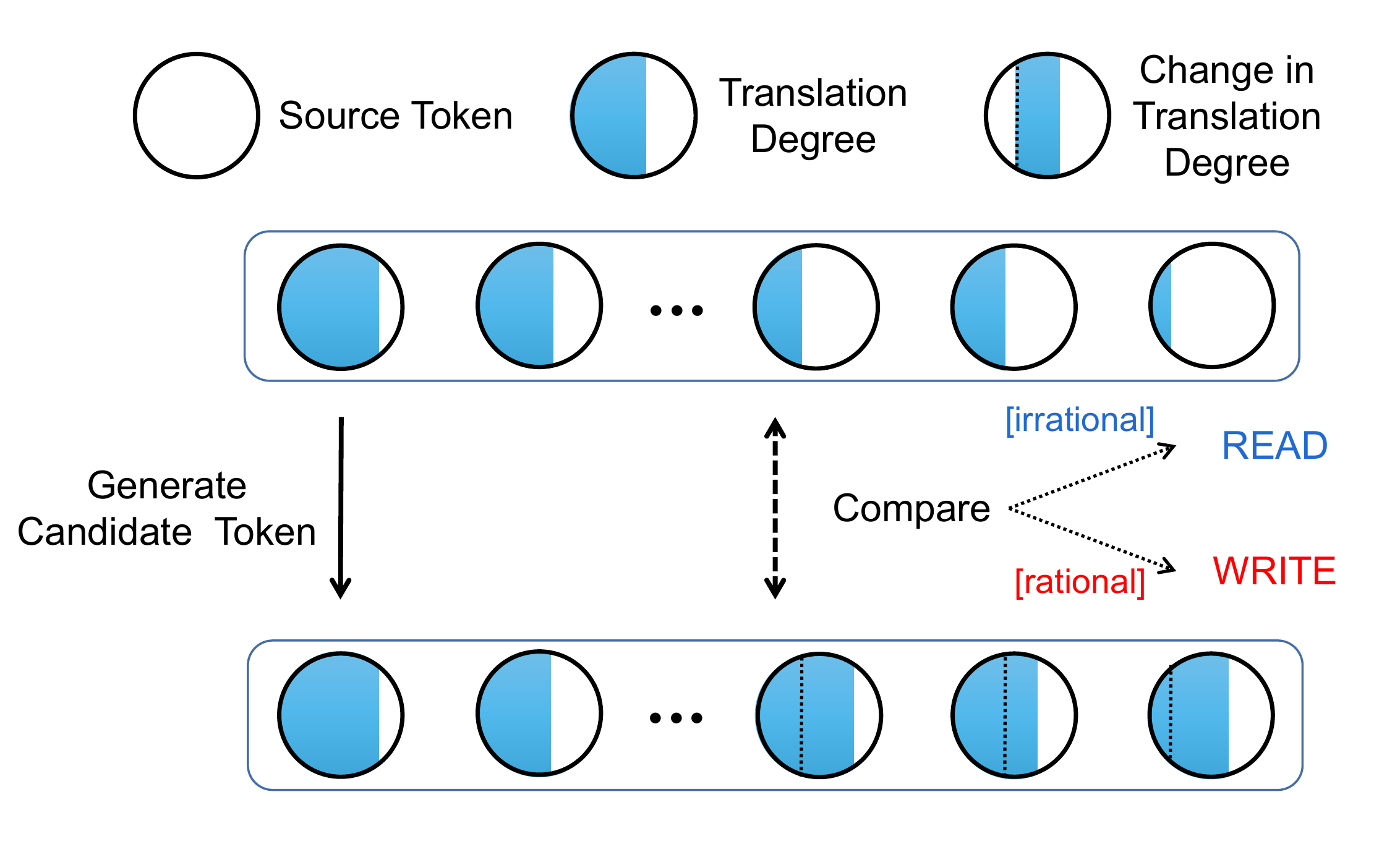}
    \caption{The change in translation degree of source tokens after generating a candidate token, and the READ/WRITE action is taken accordingly.}
    \label{comparison}
\end{figure}

However, the previous methods, including fixed and adaptive policies, lack evaluation before taking the next action. For fixed policy \citep{DBLP:conf/acl/MaHXZLZZHLLWW19, multiPath, DBLP:conf/aaai/ZhangFL21, zhang-feng-2021-universal}, the model generates translation according to the predefined translation rules. Although it only relies on simple training methods, it cannot make full use of the context to decide an appropriate translation policy. For adaptive policy \citep{gu-etal-2017-learning, arivazhagan-etal-2019-monotonic, DBLP:conf/iclr/MaPCPG20, wait-info}, the model can obtain better translation performance. But it needs complicated training methods to obtain translation policy and takes action immediately after making decisions, which usually does not guarantee the accuracy of actions.

Therefore, we attempt to explore some factors from the translation to reflect whether the action is correct, thereby introducing evaluation into translation policy. The goal of translation is to convert sentences from the source language to the target language \citep{DBLP:conf/wmt/MujadiaS21}, so the source and target sentences should contain the same semantics (i.e., \emph{global equivalence}). To ensure the faithfulness of translation \citep{DBLP:conf/emnlp/WengYWL20}, the source content that has already been translated should be semantically equivalent to the previously generated target tokens at each step (i.e., \emph{partial equivalence}) \citep{DBLP:conf/acl/ZhangF22}. Furthermore, by comparing the changes between adjacent steps, the increment of the source content being translated should be semantically equivalent to the current generated token (i.e., \emph{incremental equivalence}). Therefore, the rationality of the generated target token can be reflected by the increment of the source content being translated between adjacent steps, which can be used to evaluate the READ and WRITE actions.

In this paper, we propose a method of performing the adaptive policy by integrating \emph{post-evaluation} into the fixed policy, which directs the model to take READ or WRITE action based on the evaluation results. Using partial equivalence, our model can recognize the translation degree of source tokens (i.e., the degree to which the source token has been translated), which represents how much the source content is translated at each step. Then naturally, by virtue of incremental equivalence, the increment of translated source content can be regarded as the change in the translation degree of available source tokens. Therefore, we can evaluate the action by measuring the change in translation degree. As shown in Figure \ref{comparison}, if the translation degree has significant changes after generating a candidate token, we think that the current generated token obtains enough source content, and thus WRITE action should be taken. Otherwise, the model should continue to take READ actions to wait for the arrival of the required source tokens. Experiments on WMT15 De$\rightarrow$En and IWSLT15 En$\rightarrow$Vi translation tasks show that our method can exceed strong baselines under all latency.

\section{Background}
Transformer \cite{Transformer}, which consists of encoder and decoder, is the most widely used neural machine translation model. Given a source sentence $\mathbf{x}$ = $(x_1, ... , x_I)$, the encoder maps it into a sequence of hidden states $\mathbf{z}$ = $(z_1, ... , z_I)$. The decoder generates target hidden states $\mathbf{h}$ = $(h_1, ... , h_M)$ and predicts the target sentence $\mathbf{y}$ = $(y_1, ... , y_M)$ based on $\mathbf{z}$ autoregressively.

Our method is based on wait-$k$ policy \cite{DBLP:conf/acl/MaHXZLZZHLLWW19} and Capsule Networks \cite{capNet} with Guided Dynamic Routing \cite{zheng-etal-2019-dynamic}, so we briefly introduce them.

\subsection{Wait-$k$ Policy}
Wait-$k$ policy, which belongs to fixed policy, takes $k$ READ actions first and then takes READ and WRITE actions alternately. Define a monotonic non-decreasing function $g(t)$, which represents the number of available source tokens when translating target token $y_{t}$. For wait-$k$ policy, $g(t)$ can be calculated as:
\begin{equation}
    g(t;k) = \min\{ k \!+\! t \!-\! 1, I\} ,
\end{equation}
where $I$ is the length of the source sentence.

To avoid the recalculation of the encoder hidden states when a new source token is read, unidirectional encoder \citep{multiPath} is proposed to make each source token only attend to its previous tokens. Besides, multi-path method \citep{multiPath} optimizes the model by sampling $k$ uniformly during training and makes a unified model obtain the translation performance comparable to wait-$k$ policy under all latency.

\subsection{Capsule Networks with Guided Dynamic Routing}

Guided Dynamic Routing (GDR) is a variant of routing-by-agreement mechanism \citep{DBLP:conf/nips/Sabour} in Capsule Networks and makes input capsules route to corresponding output capsules driven by the decoding state at each step. In detail, encoder hidden states $\mathbf{z}$ are regarded as a sequence of input capsules, and a layer of output capsules is added to the top of the encoder to model different categories of source information. The decoding state then directs each input capsule to find its affiliation to each output capsule at each step, thereby solving the problem of assigning source tokens to different categories.

\section{The Proposed Method}

\begin{figure}[t]
    \centering
    \includegraphics[width=\columnwidth]{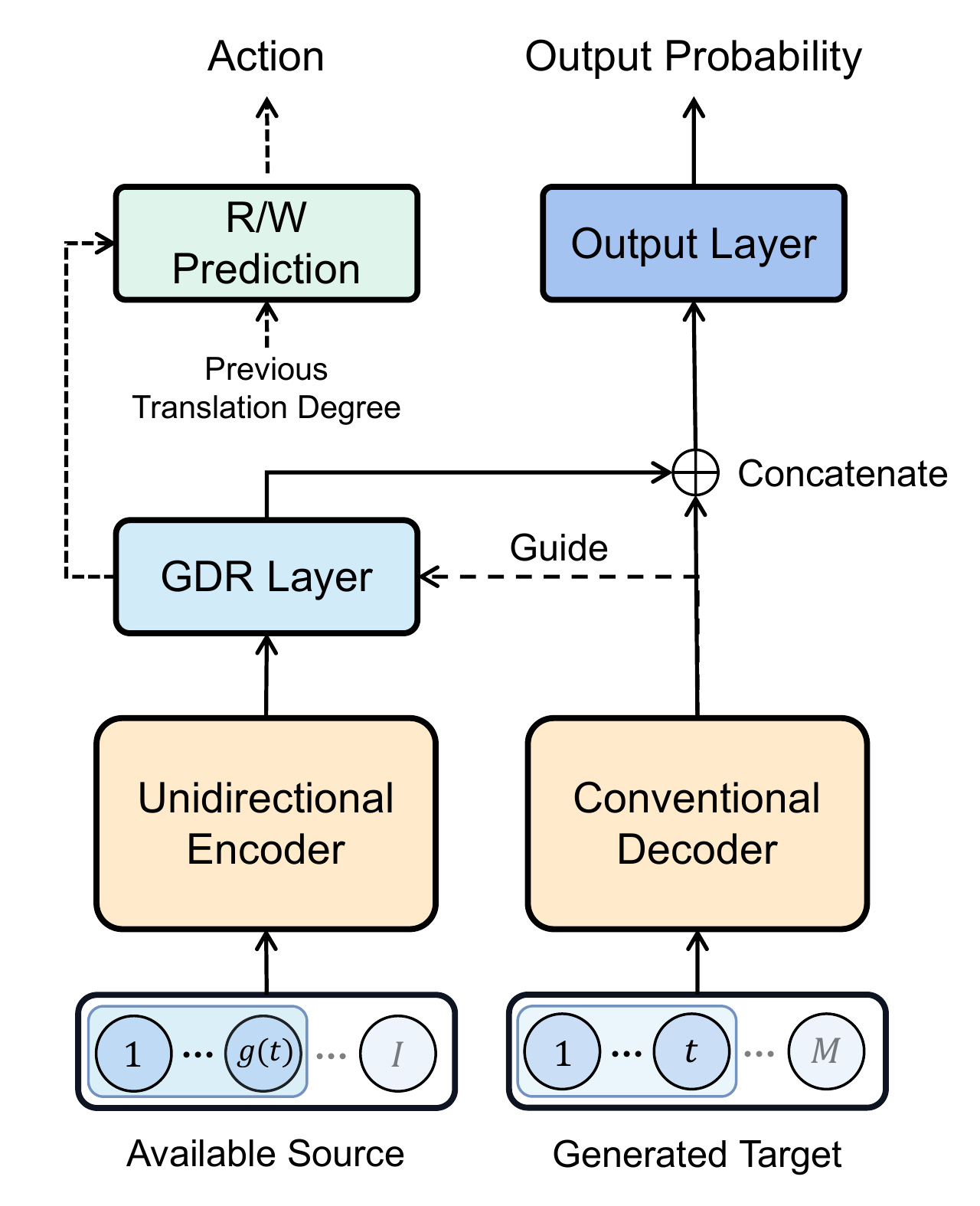}
    \caption{The architecture of our method. The R/W prediction module obtains the translation degree of the available source tokens and evaluates the next action based on the change in translation degree.}
    \label{fig-model}
\end{figure}

The architecture of our method is shown in Figure \ref{fig-model}. Our method first guides the model to recognize the translation degree of available source tokens based on partial equivalence during training via the introduced GDR module. Then based on the incremental equivalence between adjacent steps, our method utilizes the changes in translation degree to post-evaluate the rationality of the READ and WRITE actions and accordingly make corrections, thereby performing an adaptive policy during inference. Besides, to enhance the robustness of the model in recognizing the translation degree during inference, our method applies a disturbed-path training based on the wait-$k$ policy, which adds some disturbance to the translation policy during training. The details are introduced in the following sections in order.

\subsection{Recognizing the Translation Degree}
\label{lossFunction}
As mentioned above, the translation degree represents the degree to which the source token has been translated and is the prerequisite of our method. Therefore, we introduce Capsule Networks with GDR to model the translation degree, which is guided by our proposed two constraints according to partial equivalence during training.

\paragraph{Translation Degree}
We define the translation degree of all source tokens at step $t$ as $\mathbf{d}^{(t)}$ = $(d^{(t)}_1, ... ,d^{(t)}_I)$. To obtain the translation degree, we need to utilize the ability of Capsule Networks with GDR to assign the source tokens to different categories. Assume that there are $J\!+\!N$ output capsules modeling available source information that has already been translated and has not yet been translated, among which there are $J$ translated capsules $\mathbf{\Phi}^T$ = $(\Phi_1, ... , \Phi_{J})$ and $N$ untranslated capsules $\mathbf{\Phi}^U$ = $(\Phi_{J\!+\!1}, ... , \Phi_{J\!+\!N})$, respectively. The encoder hidden states $\mathbf{z}$ are regarded as input capsules. To determine how much of $z_i$ needs to be sent to $\Phi_j$ at step $t$, the assignment probability $c_{ij}^{(t)}$ in SiMT is modified as:
\begin{equation}
  c_{ij}^{(t)} = \left\{\begin{matrix}  
  \frac{\exp b_{ij}^{(t)}}{\sum_{l} \exp b_{il}^{(t)}}  & \text{if} \;\; i \leq g(t)  \\[0.05cm]
  0 &\text{otherwise}
  \end{matrix}\right. ,
\label{assign_prob}
\end{equation}
where $b_{ij}^{(t)}$ measures the cumulative similarity between $z_i$ and $\Phi_j$. 
Then $c_{ij}^{(t)}$ is updated iteratively driven by the decoding state and is seen as the affiliation of $z_i$ belonging to $\Phi_j$ after the last iteration. For more details about Capsule Networks with GDR, please refer to \citet{zheng-etal-2019-dynamic}. On this basis, the translation degree of $x_i$ is calculated by aggregating the assignment probability of routing to the translated capsules at step $t$:
\begin{equation}
    d_i^{(t)} = \sum\limits_{j = 1}^{J} c^{(t)}_{ij}.
    \label{translationD}
\end{equation}

\paragraph{Segment Constraint}
To ensure that the model can recognize the translation degree of source tokens, the model requires additional guidance. According to partial equivalence, the translated source content should be semantically equivalent to the generated target tokens. On the contrary, the untranslated source content and unread source tokens should be semantically equivalent to target tokens not generated. So we introduce mean square error to induce the learning of output capsules:
\begin{equation}
\begin{split}
\mathcal{L}_\textsc{S}   &= \frac{1}{M} \sum\limits_{t = 1}^{M}({\Vert \mathbf{\Phi}^T_{t} - {\mathbf{W}^T} \mathbf{H}^T_t \Vert}^2 \\ &+ {\Vert \mathbf{\Phi}^U_{t} + {\mathbf{W}^U_e} \mathbf{Z}_t - {\mathbf{W}^U_d} \mathbf{H}^U_t \Vert}^2 )
\end{split} ,
\end{equation}
where $\mathbf{W}^T$, $\mathbf{W}^U_e$ and $\mathbf{W}^U_d$ are learnable parameters. $\mathbf{H}^T_t$ and $\mathbf{H}^U_t$ are the averages of hidden states of the generated target tokens and target tokens not generated, which are calculated respectively:
\begin{equation}
\mathbf{H}^T_t = \frac{1}{t\!-\!1} \sum\limits_{\tau = 1}^{t-1} h_\tau ,
\end{equation}
\begin{equation}
\mathbf{H}^U_t = \frac{1}{M\!-\!t+1} \sum\limits_{\tau = t}^{M} h_\tau .
\end{equation}
where $M$ is the length of the target sentence. $\mathbf{Z}_t$ is the average of hidden states of unread source tokens at step $t$:
\begin{equation}
\mathbf{Z}_t = \frac{1}{I\!-\!g(t)} \sum\limits_{\tau = g(t)+1}^{I} z_\tau .
\end{equation}
$\mathbf{\Phi}^T_{t}$ and $\mathbf{\Phi}^U_{t}$ are the translated and untranslated source information at step $t$, respectively.

\paragraph{Token Constraint}

To recognize the changes in translation degree more accurately, we propose token constraint according to incremental equivalence. It encourages the translated capsules to predict the generated tokens and combines translated and untranslated capsules to predict the available source tokens at each step. It can be calculated as:
\begin{equation}
\begin{split}
\mathcal{L}_\textsc{T} &= - \frac{1}{M} \sum\limits_{t = 1}^{M}[\log p_d(\mathbf{y}_{<t}|\mathbf{\Phi}^T_t) \\
&+ \log p_e(\mathbf{x}_{\leq g(t)}|\mathbf{\Phi}^T_t;\mathbf{\Phi}^U_t)]
\end{split} ,
\end{equation}
where $p_d(\mathbf{y}_{<t}|\mathbf{\Phi}^T_t)$ represents the probability of generated target tokens based on translated source information and $p_e(\mathbf{x}_{\leq g(t)}|\mathbf{\Phi}^T_t;\mathbf{\Phi}^U_t)$ is the probability of available source tokens based on both translated and untranslated information. Then we can get the training objective of our model:
\begin{equation}
    \mathcal{L}(\theta) = \!-\!\log p_{\theta}(\mathbf{y}|\mathbf{x}) + \lambda_{S}\mathcal{L}_{S} + \lambda_{T}\mathcal{L}_{T} ,
    \label{lossTerm}
\end{equation}
where$\,$ $\!-\!\log p_{\theta}(\mathbf{y}|\mathbf{x})$ is negative log-likelihood.

\begin{figure*}[t]
\centering
\subfigure[WRITE Action]{
\includegraphics[width=2.95in]{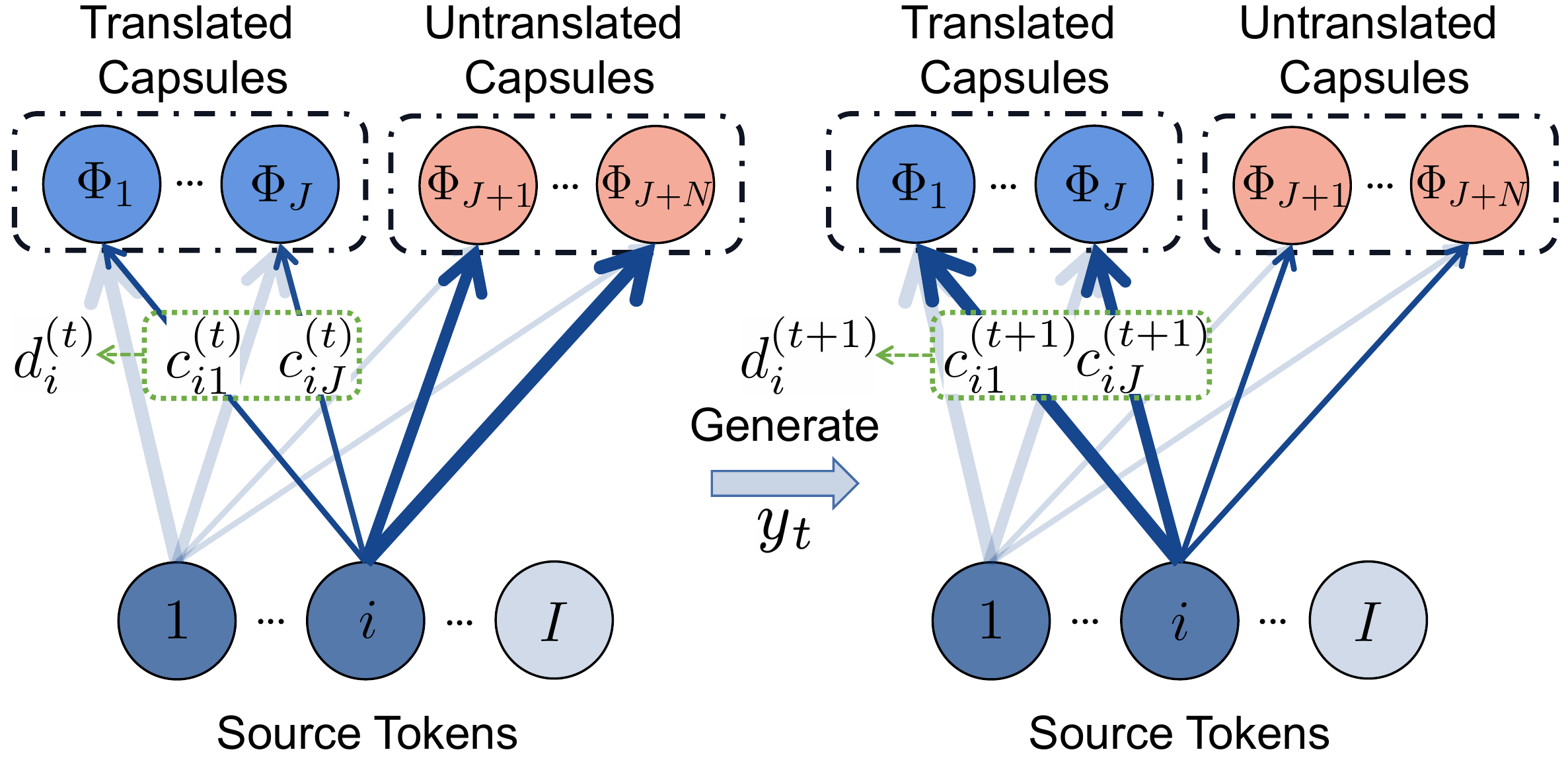}
}
\quad
\subfigure[READ Action]{
\includegraphics[width=2.95in]{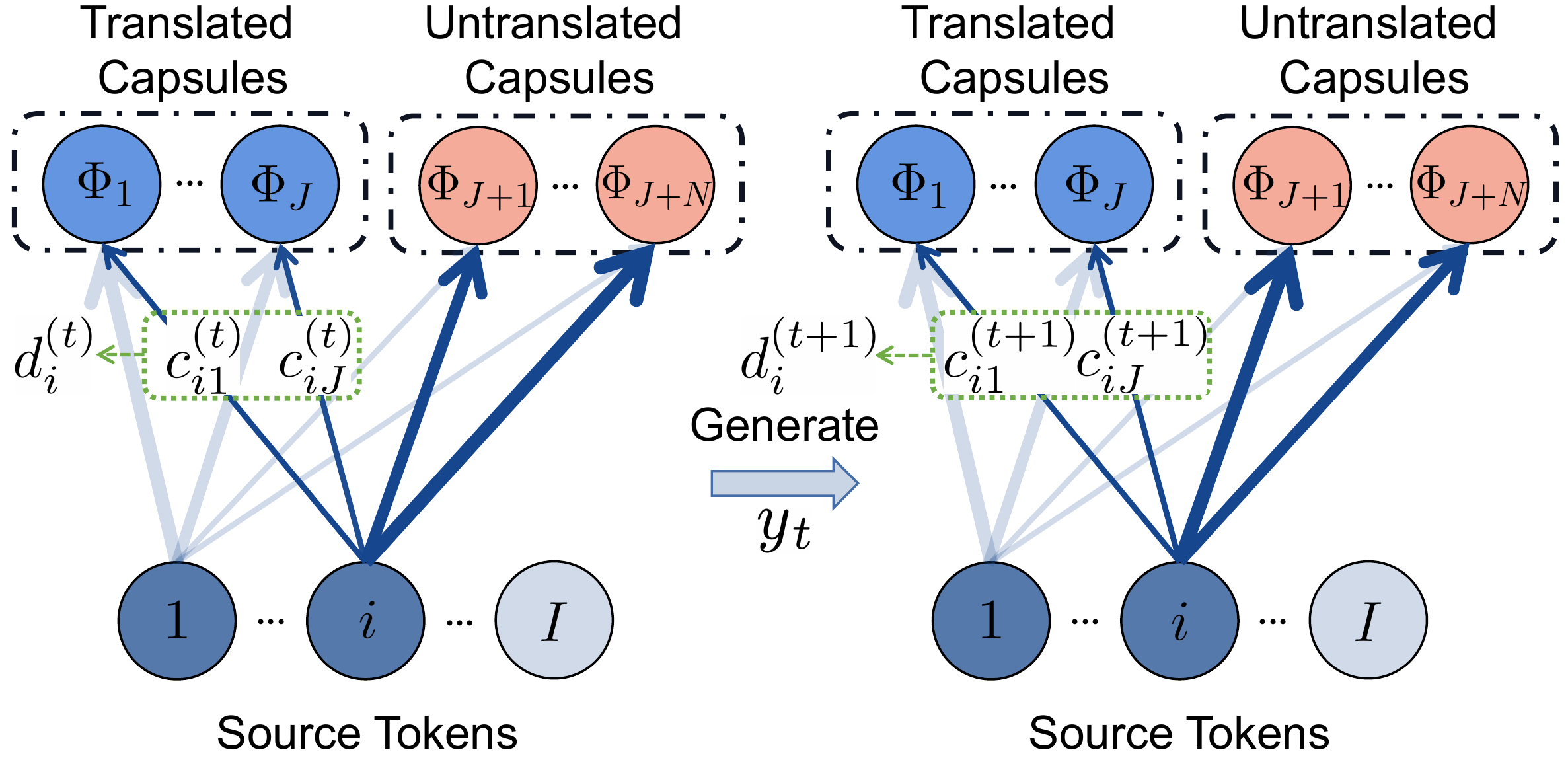}
}

\caption{Change in translation degree of available sources token after generating $y_t$. The model takes WRITE action when the translation degree has significant changes. Otherwise, the model should take READ action.}
\label{RWOP}
\end{figure*}
\subsection{Post-Evaluation Policy}
With the help of token and segment constraints, our model can accurately recognize the translation degree, which can be utilized to perform our Post-Evaluation (PE) policy by measuring the changes in translation degree between adjacent steps.

Generally speaking, the core of the adaptive policy is to decide the conditions for taking different actions \citep{ITST}. According to incremental equivalence, the current generated token should be semantically equivalent to the increment of the source content that has been translated, which can be measured by the changes in translation degree. Therefore, we can evaluate the rationality of actions by measuring the change in the translation degree of available source tokens. We define the change in the translation degree of source tokens after generating $y_t$ as $\Delta \mathbf{d}^{(t)}$ = $(\Delta d^{(t)}_1, ... , \Delta d^{(t)}_I)$ and $\Delta d^{(t)}_i$ is calculated as:
\begin{equation}
    \Delta d_i^{(t)} = \max \{d_i^{(t+1)} \!-\! d_i^{(t)}, 0\} ,
    \label{diff}
\end{equation}
where $d^{(t)}_i$ and $d^{(t+1)}_i$ are calculated in Eq.(\ref{translationD}) and $\max(\cdot)$ function ensures that the translation degree is undiminished considering incremental equivalence. Furthermore, we introduce hyperparameter $\rho$, which is the threshold to measure the change in translation degree.

As shown in Figure \ref{RWOP}, we can get the conditions for taking different actions by comparing $\Delta \mathbf{d}^{(t)}$ and $\rho$. We first define function $\mathrm{max\_select(\cdot)}$, which returns the maximum element in a vector. According to incremental equivalence, if the change in the translation degree exceeds the threshold (i.e, $\mathrm{max\_select(\Delta \mathbf{d}^{(t)})} \!\geq\! \rho$), then the current generated token obtains enough source content, and the model should take WRITE action. Otherwise, the model should continue to take READ action. However, the generation of auxiliary tokens such as `the' in English can not lead to a change in translation degree. This misleads the model to take READ actions consecutively, so we force the model to take WRITE actions by setting the restriction of consecutive READ actions as $r$. PE policy is shown in Algorithm \ref{algor}. Our model will only take WRITE action after reading the whole source sentence.

\begin{algorithm}[t]
  \SetAlgoLined
  \SetKwProg{Fn}{Function}{:}{end}
  \SetKwProg{ret}{return}{}{}
  \SetKwFunction{Eva}{Evaluation}
  \SetKwInput{KwData}{Data}
  \KwIn{Threshold $\rho$, Restriction on READ actions $r$, $y_0 \leftarrow \left \langle bos \right \rangle$, Prefix with $k$ source tokens $\bm{\mathrm{x}}_{\leq k}$, $t \leftarrow 1$, $i \leftarrow k$}
  \While{$y_{t-1} \neq \left \langle eos \right \rangle$}{
    
    \uIf{\Eva{$\bm{\mathrm{x}}_{\leq i}$, $\bm{\mathrm{y}}_{< t}$, $\rho$}}{                     
      Take \textbf{WRITE} action
      
      $t \leftarrow t+1$
      }\uElse{
      Take \textbf{READ} action
      
      $i \leftarrow i+1$
      }
    }
  \Fn(){\Eva{$\bm{\mathrm{x}}_{\leq i}$, $\bm{\mathrm{y}}_{<t}$, $\rho$}}{
    calculate $\mathbf{d}^{(t)}$ as Eq.(\ref{translationD})
    
    generate $y_t$ \tcp*[f]{$\!\!\!\triangleright $Candidate}
    
    calculate $\mathbf{d}^{(t+1)}$ as Eq.(\ref{translationD})
    
    calculate $\Delta \mathbf{d}^{(t)}$ as Eq.(\ref{diff})
    
    \uIf{$\mathrm{max\_select{(\Delta \mathbf{d}^{(t)}})} \geq \rho$}{
        \ret(True){}{}
    }\uElse{
        \ret(False){}{}
    }
  }
  \caption{Post-Evaluation Policy}
  \label{algor}
\end{algorithm}

\subsection{Disturbed-Path Training}
Up to now, we have proposed our adaptive policy by introducing post-evaluation, which utilizes the translation degree. Because the adaptive policy adopts different translation paths (i.e., the sequence of READ and WRITE actions) for different contexts, this requires the model to learn as many translation paths as possible. However, the previous training methods \citep{DBLP:conf/acl/MaHXZLZZHLLWW19, multiPath} can only cover a small number of predefined translation paths. To enhance the ability to recognize the translation degree on different translation paths, our model is optimized across our proposed disturbed-path.

Specifically, the log-likelihood estimation based on sentence pair ($\mathbf{x}$, $\mathbf{y}$) through the single path $\mathbf{g}_k$ is computed as:
\begin{equation}
    \log p(\mathbf{y}|\mathbf{x}, \mathbf{g}_k) = \sum\limits_{t = 1}^{M} \log p(y_t|\mathbf{y}_{<t}, \mathbf{x}_{\leq g(t;k)}) ,
    \label{pre_opt}
\end{equation}
where $\mathbf{g}_k$ = $(g(1;k), ..., g(M;k))$ defines the number of available source tokens at each step and $k$ is the number of source tokens read in advance before generation. For translation path $\mathbf{g}_k$, $g(t;k)$ is updated as:
\begin{equation}
  g(t;k) = \left\{\begin{matrix}
  \min\{g(t-1;k)+\gamma, I\},&t>1\\
  \min\{k+\gamma, I\},& t=1
  \end{matrix}\right. ,
\label{eq:simulsoft}
\end{equation}
where $\gamma$ is uniformly sampled from $[0,...,r]$ and $r$ is the restriction on READ actions in PE policy and controls the degree of disturbance to a single translation path. This essentially simulates the situation where the model makes decisions on the next action. For ($\mathbf{x}$, $\mathbf{y}$), we then make the model have the ability to recognize the translation degree under all latency by changing $k$. Thus, the log-likelihood estimation in Eq.(\ref{pre_opt}) is modified:
\begin{equation}
    E_k[\log p(\mathbf{y}|\mathbf{x}, \mathbf{g}_k)] = \sum\limits_{k \sim \mathcal{U}(\emph{\rm{K}})} \log p(\mathbf{y}|\mathbf{x}, \mathbf{g}_k) ,
\end{equation}
where $k$ is uniformly sampled form \emph{\rm{K}} = $ [1,...,I]$ and $I$ is the length of source sentence. Therefore, our method can perform our adaptive policy under all latency by only using a unified model.

\section{Experiments}

\subsection{Datasets}
We evaluate our proposed method on IWSLT15\footnote{\url{https://nlp.stanford.edu/projects/nmt/}} English$\rightarrow$Vietnamese (En$\rightarrow$Vi) task, IWSLT14\footnote{\url{https://wit3.fbk.eu/2014-01}} English$\rightarrow$German (En$\rightarrow$De) task, and WMT15\footnote{\url{www.statmt.org/wmt15/}} German$\rightarrow$English (De$\rightarrow$En) task.

For En$\rightarrow$Vi task \cite{DBLP:conf/iwslt/CettoloNSBCF16}, our settings are the same as \citet{arivazhagan-etal-2019-monotonic}. We replace tokens whose frequency is less than 5 with $\left \langle unk \right \rangle$. We use TED tst2012 as the development set and TED tst2013 as the test set.

For En$\rightarrow$De task, the model settings remain the same as \citet{DBLP:conf/iwslt/CettoloNSBF14}.

For De$\rightarrow$En task, we keep our settings consistent with \citet{DBLP:conf/iclr/MaPCPG20}. We apply BPE \cite{sennrich-etal-2016-neural} with 32K subword units and use a shared vocabulary between source and target. We use newstest2013 as the development set and newstest2015 as the test set.

\begin{figure*}[t]
\centering
\subfigure[En$\rightarrow$Vi]{
\includegraphics[width=0.32\textwidth]{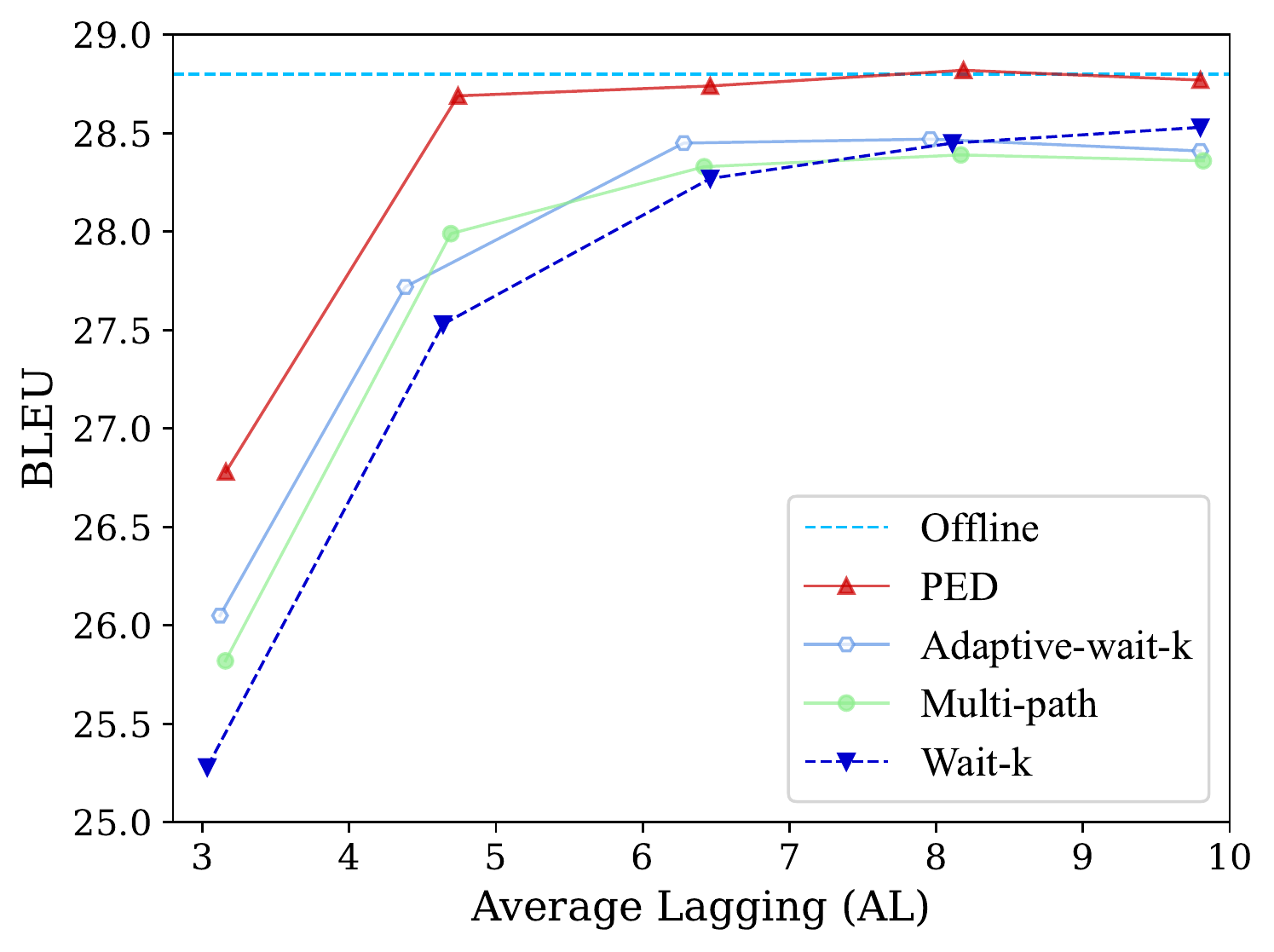}
}
\subfigure[En$\rightarrow$De]{
\includegraphics[width=0.31\textwidth]{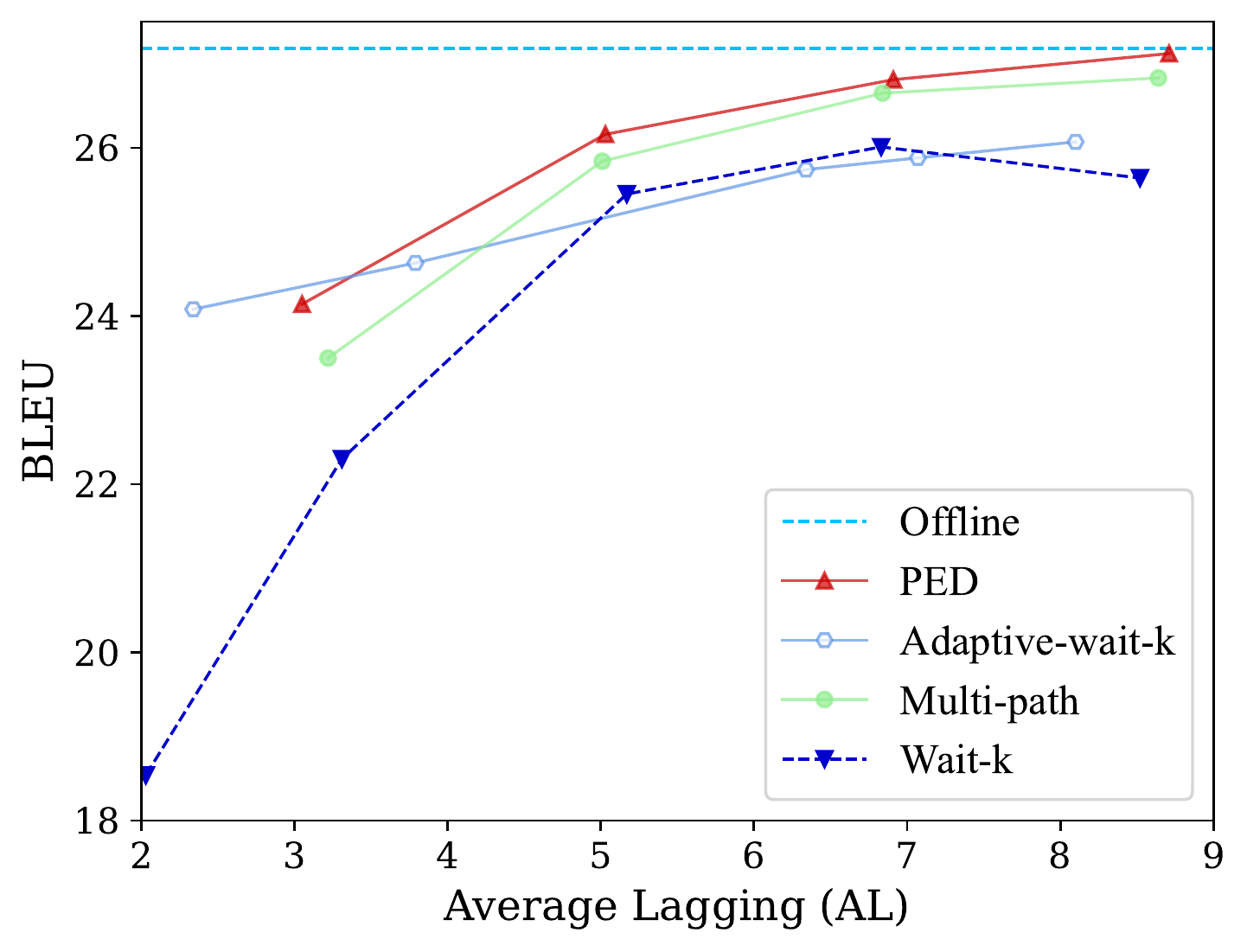}
}
\subfigure[De$\rightarrow$En]{
\includegraphics[width=0.31\textwidth]{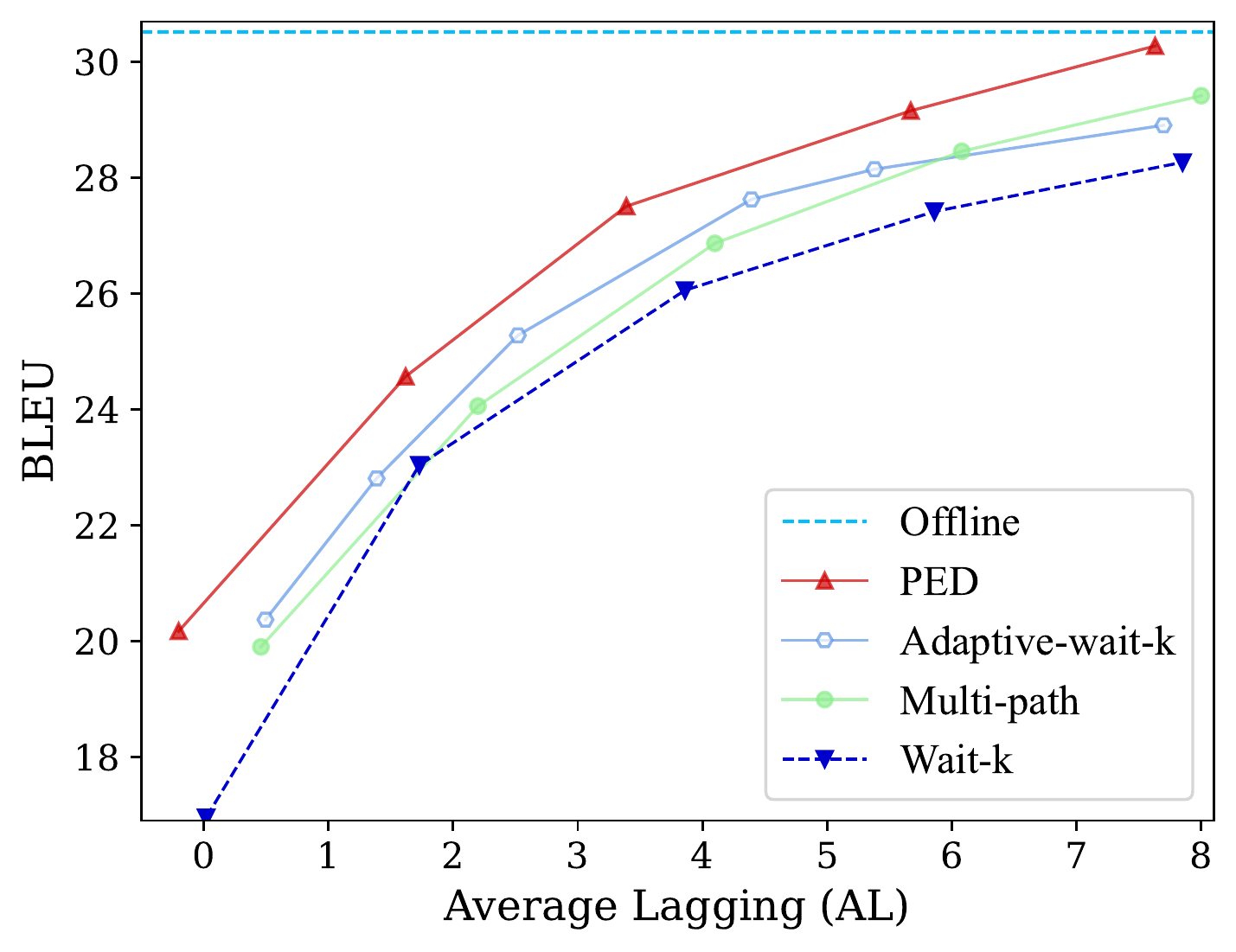}
}
\caption{Performance of different methods on En$\rightarrow$Vi (Transformer-Small), En$\rightarrow$De (Transformer-Small) and De$\rightarrow$En (Transformer-Base) tasks. It shows the results of our methods, wait-$k$, multi-path, adaptive-wait-$k$ and offline model.}
\label{mainRes}
\end{figure*}

\subsection{Model Settings}
\label{modelSetting}
Since our experiments involve the following models, we briefly introduce them. \textbf{Wait-$k$} \citep{DBLP:conf/acl/MaHXZLZZHLLWW19} policy is the benchmark method in SiMT. It takes $k$ READ actions first, and then alternates between READ and WRITE actions. \textbf{Multi-path} \citep{multiPath} achieves comparable performance to wait-$k$ policy under all latency with a unified model. \textbf{Adaptive-wait-$k$} \citep{DBLP:conf/acl/ZhengLZMLH20} implements the adaptive policy through a heuristic composition of several fixed policies. \textbf{Offline} refers to conventional Transformer \citep{Transformer} for full-sentence machine translation. \textbf{PED} represents that our model is trained through disturbed-path and performs PE policy during inference. For all the models mentioned above, we apply Transformer-Small (6 layers, 4 heads) on En$\rightarrow$Vi and En$\rightarrow$De tasks and Transformer-Base (6 layers, 8 heads) on De$\rightarrow$En task. Other model settings follow \citet{DBLP:conf/iclr/MaPCPG20}.

We implement all models by adapting Transformer from Fairseq Library \cite{DBLP:conf/naacl/OttEBFGNGA19}. The settings of Capsule Networks with GDR are consistent with \citet{zheng-etal-2019-dynamic}. For our method, we empirically set $r$ = 2 and $\rho$ = 0.24 for all experiments, and use $k$ as free parameter to achieve different latency. Our proposed method is fine-tuned based on the pre-trained multi-path model. We use greedy search in decoding and evaluate these methods with translation quality measured by tokenized BLEU \citep{papineni-etal-2002-bleu} and latency estimated by Average Lagging (AL) \citep{DBLP:conf/acl/MaHXZLZZHLLWW19}.

\subsection{Main Results}
The translation performance between our method and the previous methods is shown in Figure \ref{mainRes}. It can be seen that our method can exceed previous methods under all latency on all translation tasks.

Compared to wait-$k$ policy, our method obtains significant improvement, especially under low latency. This is because wait-$k$ policy performs translation according to the predefined path, which usually leads to uncertain anticipation or introduces redundant latency \citep{DBLP:conf/acl/MaHXZLZZHLLWW19}. Both Multi-path and our methods can generate translation under all latency with a unified model. But our PED method transcends its performance by performing Post-Evaluation (PE) policy, which can evaluate the rationality of actions and then decide whether to take them. Therefore, compared with fixed policy, our PE method can achieve better performance by adjusting its translation policy.

Compared to Adaptive-wait-$k$ policy, our model also surpasses its performance and is more reliable under high latency. Adaptive-wait-$k$ generates translation through a heuristic composition of several models with different fixed policies, which restricts the performance under high latency and leads to a decrease in translation speed caused by frequent model switching \citep{DBLP:conf/acl/ZhengLZMLH20}. Our method generates translation with only a unified model and integrates post-evaluation into fixed policy to evaluate the rationality of actions. In particular, our model can approach the performance of full-sentence machine translation with lower latency on two tasks.

\begin{figure*}[t]
\centering
\includegraphics[width=6.3in]{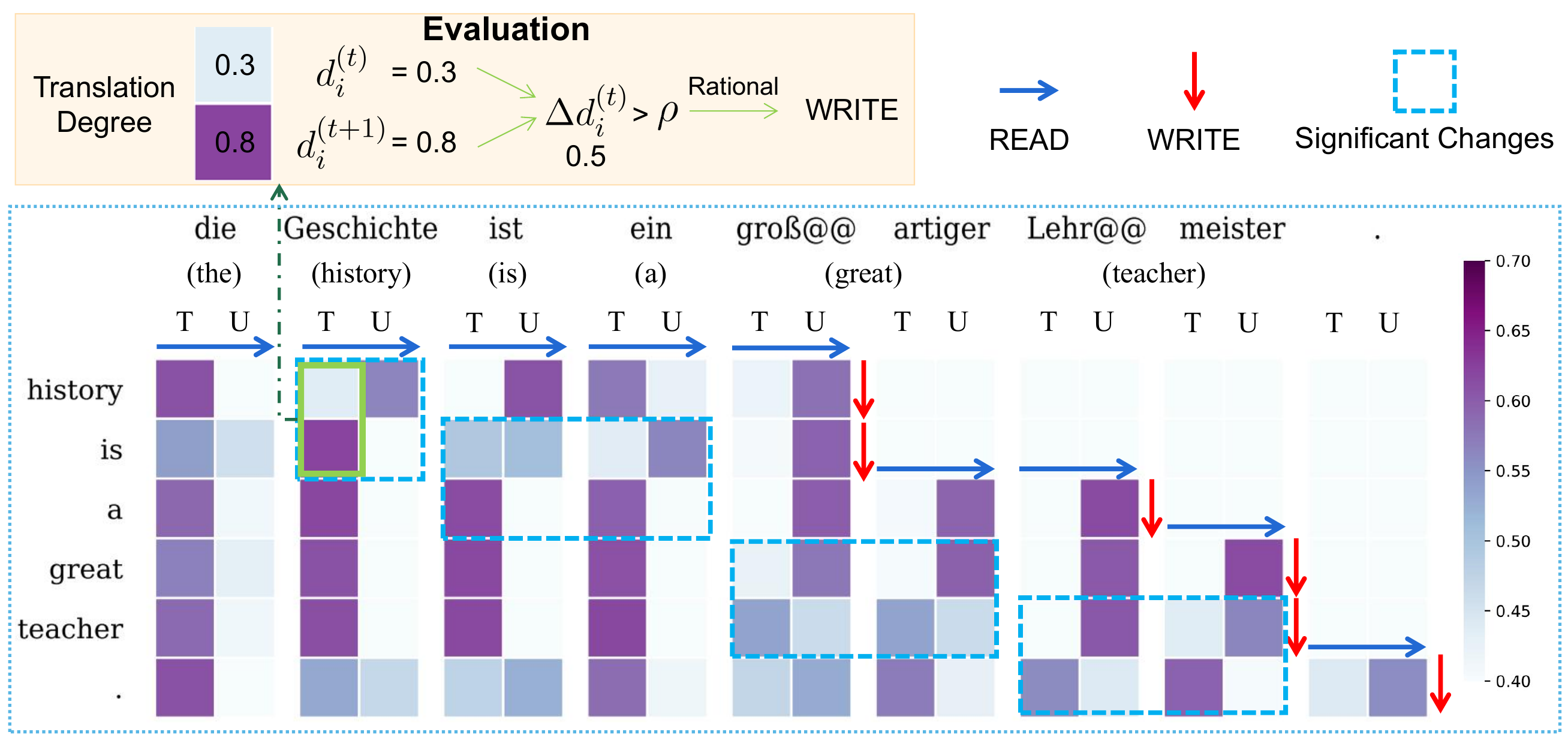}
\caption{Translation and Evaluation process of a De$\rightarrow$En example when performing PE policy with $k$ = 5. The horizontal direction denotes the source sentence (De), and the vertical direction denotes generated sentence (En). `T' represents the translation degree. `U' represents the degree to which the source token has not yet been translated. Our PE policy can take WRITE actions accurately when the translation degree has significant changes.}
\label{vis}
\end{figure*}

\section{Analysis}
To understand our proposed method, we conduct multiple analyses. All of the following results are reported on De$\rightarrow$En task.

\begin{table}[t]
\normalsize
\centering
\begin{tabular} {l | c | c}
\hline
 & AL & BLEU \\
\hline
PED  & 7.63 & 30.28 \\
\hline
$\;\;$w/o PE & 7.9 & 30.10 \\
$\;\;$w/o disturbed-path & 7.81 & 29.68 \\
$\;\;$w/o PE, disturbed-path & 7.59 &  29.48\\
\hline
\end{tabular}
\caption{Ablation study of our method when $k$ = 9. `w/o PE' denotes our model is trained across disturbed-path and performs fixed policy. `w/o disturbed-path' denotes our model is trained across multi-path and performs our PE policy.
}
\label{Components}
\end{table}

\subsection{Ablation Study}
We conduct an ablation study on PE policy and disturbed-path training method to verify their effectiveness, respectively. As shown in Table \ref{Components}, both PE policy and disturbed-path method can improve the translation performance, and better latency-quality tradeoffs can be obtained by their joint contributions.

We also carry out comparative experiments to understand the two constraints in subsection \ref{lossFunction}. The results are shown in Table \ref{supervision_signal}. Both token and segment constraints have positive effects on translation performance respectively. Although the translation quality is slightly worse when the model is guided by them concurrently, the translation degree of available source tokens can be greatly improved and the latency is also reduced by their combined contributions.

\begin{table}[t]
\normalsize
\centering
\begin{tabular}{c|c|c|c}
    \hline
    {\;\;$\mathcal{L}_\textsc{T}$\;\;}	&	{\;\;$\mathcal{L}_\textsc{S}$\;\;}	&	{\;\;AL\;\;}	&	{BLEU}\\
    \hline
    \texttimes	&	\texttimes	&	7.77		&	29.48\\
    \checkmark	&	\texttimes	&	7.86		&	29.57\\
    \texttimes	&	\checkmark	&	7.78		&	29.73\\
    \checkmark	&	\checkmark	&	7.59		&	29.48\\
    \hline
\end{tabular}
\caption{Comparison among the combinations of two constraints when decoding with $k$ = 9. The model is optimized through multi-path and performs fixed policy.
}
\label{supervision_signal}
\end{table}

\subsection{Analysis of Translation Degree}
\label{TranDegree}
To describe the translation degree intuitively, we visualize it in Figure \ref{vis}. Obviously, the translation degree of each source token gradually accumulates with the progress of translation, which means that the source content is gradually utilized by the target to generate translation and observes partial equivalence. Besides, our PE policy can take WRITE actions when the translation degree of source tokens has significant changes, which obeys incremental equivalence and ensures the rationality of actions. Therefore, our PED policy can adaptively adjust the translation path based on context to achieve better translation performance.

Following \citet{zheng-etal-2019-dynamic}, we evaluate the accuracy of the translation degree at each step by using overlapping rate, which measures the coincidence between the predicted tokens and ground-truth tokens. We introduce the prediction function in token constraint to predict the target and source tokens respectively. Then we obtain target overlapping rate $R^T$ by comparing the predicted target tokens with the generated tokens and source overlapping rate $R^S$ by comparing the predicted source tokens with available source tokens. $R^T$ is calculated as:
\begin{equation*}
    R^T = \frac{1}{M} \sum_{t=1}^M \frac{|\mathrm{Top_7}(p_d(\Phi^T_t)) \cap \mathbf{y}_{<t}|}{|\mathbf{y}_{<t}|} , 
\end{equation*}
where $p_d(\cdot)$ in subsection \ref{lossFunction} predicts the target tokens based on translated capsules and $\mathrm{Top_7}(\cdot)$ obtains $7$ tokens ($7$ is just half of the average length of the target sentence in test set) with the highest probability. $R^T$ measures the ability of translated capsules to express target information. Similarly, $R^S$ is calculated as:
\begin{equation*}
    R^S = \frac{1}{M} \sum_{t=1}^M \frac{|\mathrm{Top_{15}}(p_e(\Phi^T_t;\Phi^U_t)) \cap \mathbf{x}_{\leq g(t)}|}{|\mathbf{x}_{\leq g(t)}|}, 
\end{equation*}
where $p_e(\cdot)$ in subsection \ref{lossFunction} to predict the source tokens based on output capsules. $\mathrm{Top_{15}}(\cdot)$ obtains $15$ tokens ($15$ is just the average length of the source sentence in test set) with the highest probability. $R^S$ measures the ability of output capsules to express available source information. The results are shown in Table \ref{Similar}. The output capsules can well represent the available source information and generated target information under all latency. Therefore, our method can recognize the translation degree accurately at each step according to 
partial equivalence, thereby providing the basis for our policy.

\begin{table}[t]
\renewcommand{\arraystretch}{1.2}
\normalsize
\centering
\begin{tabular} {c | c c  c  c c}
\hline
Latency & 1 & 3 & 5 & 7 & 9\\
\hline
$R^T (\uparrow)$ & 0.60 & 0.62 & 0.63 & 0.62 & 0.61 \\
\hline
$R^S (\uparrow)$ & 0.80 & 0.78 & 0.77 & 0.77 & 0.78  \\
\hline
\end{tabular}
\caption{The results of overlapping rate under all latency, where the higher rate is better. The model is trained across disturb-path and performs fixed policy.}
\label{Similar}
\end{table}

\subsection{Analysis on Translation Path}
The purpose of the translation policy is to get a better translation path, which is composed of READ and WRITE actions. To verify the effectiveness of our PE policy, we introduce sufficiency and necessity \citep{DBLP:conf/acl/ZhangF22} as evaluation metrics. Essentially, sufficiency measures the faithfulness of the generated translation and necessity measures how much the redundant delay is introduced.

\begin{figure}[t]
\centering
\subfigure[Sufficiency]{
\includegraphics[width=2.8in]{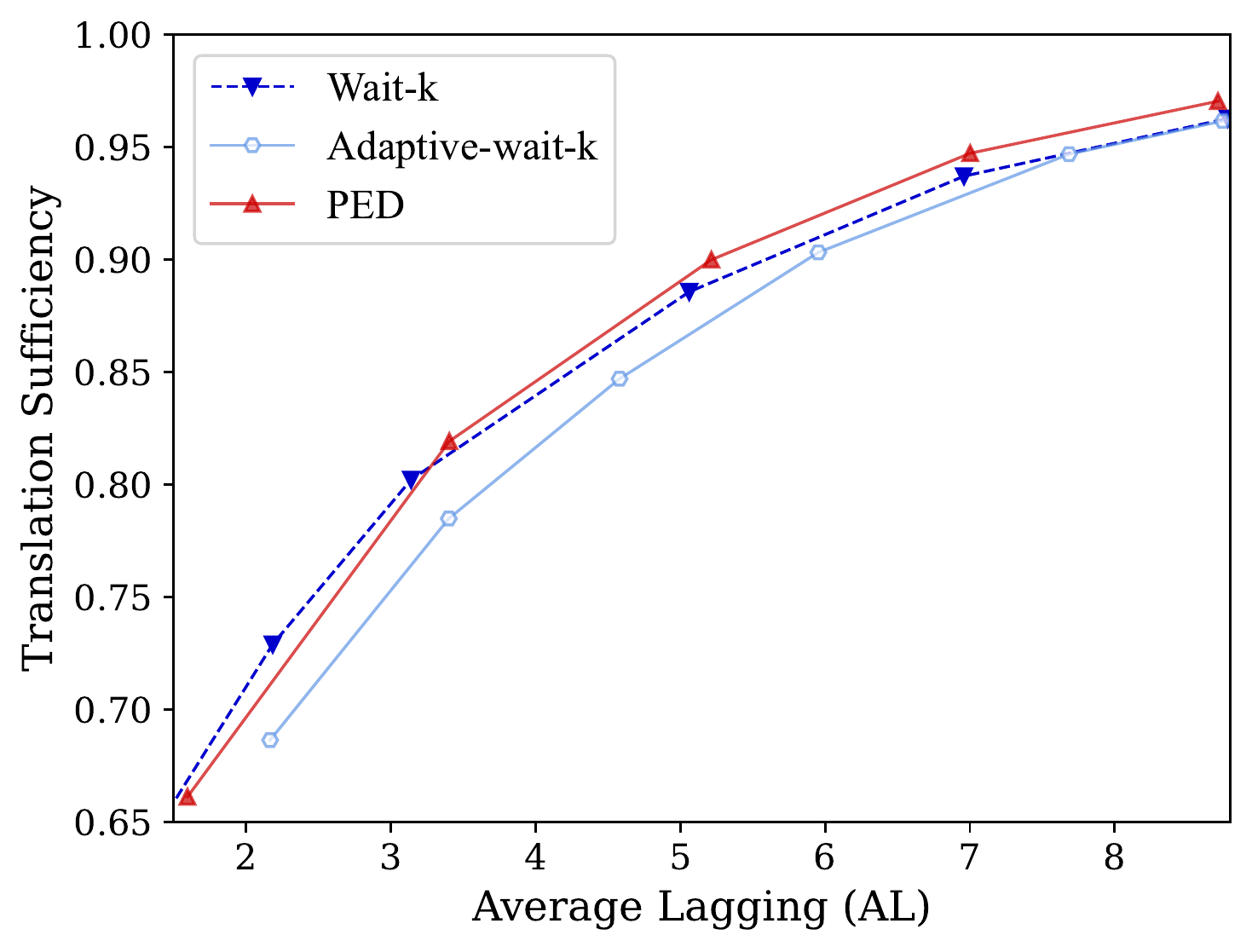}
}
\subfigure[Necessity]{
\includegraphics[width=2.8in]{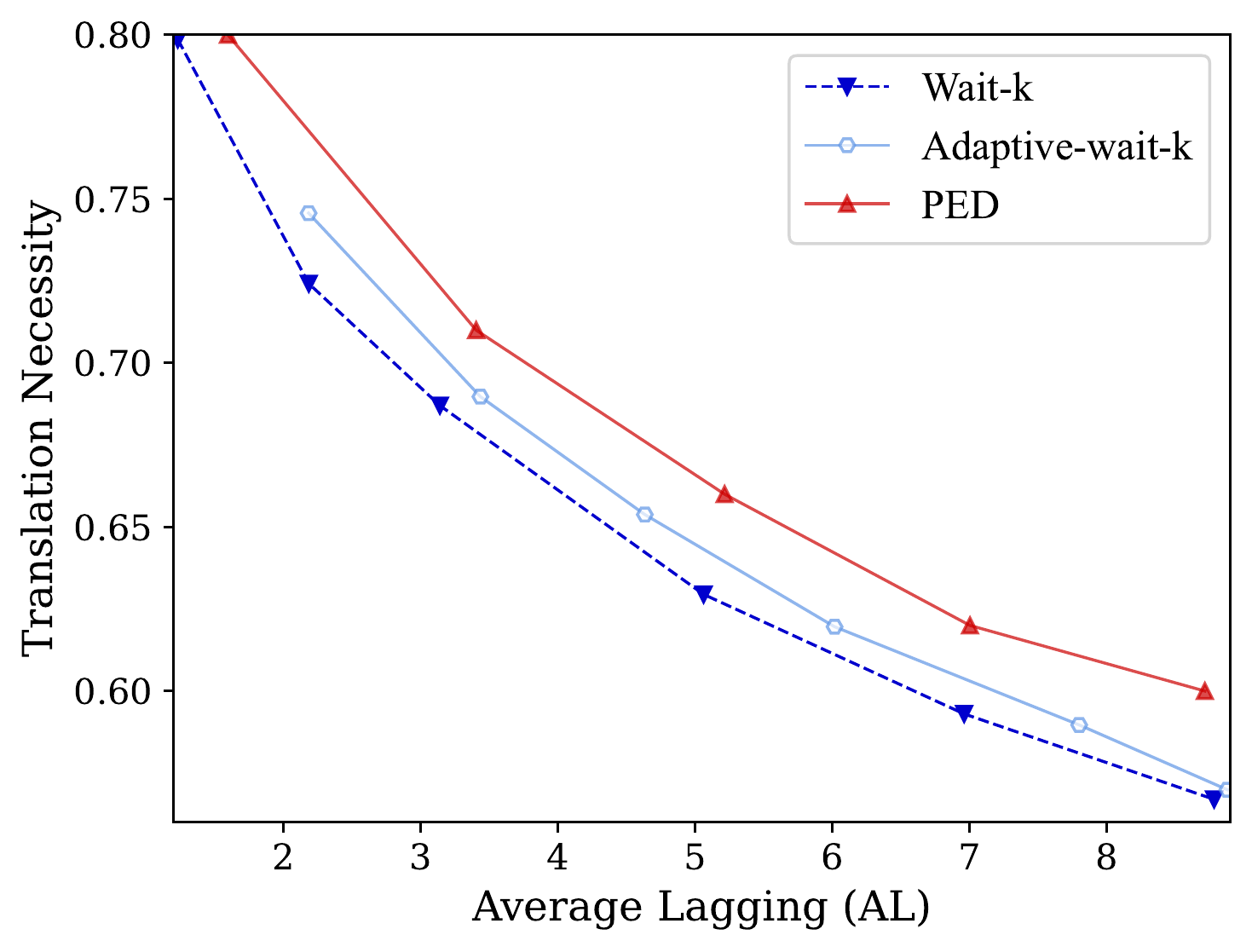}
}
\caption{Comparison of adequacy and necessity of translation path between different translation policies.}
\label{suff_nece}
\end{figure}

We take manually aligned alignments for De$\rightarrow$En corpus in RWTH dataset\footnote{\url{https://www-i6.informatik.rwth-aachen.de/goldAlignment/}} as ground-truth alignments. The comparison of sufficiency and necessity of different methods is shown in Figure \ref{suff_nece}. Obviously, the translation path decided by our PE policy exceeds other methods in terms of sufficiency and necessity. The sufficiency of wait-$k$ policy is similar to PE policy, but it introduces too much unnecessary delay under all latency. Compared to wait-$k$ policy, Adaptive-wait-$k$ performs better in terms of necessity, but it is obtained at the cost of partial sufficiency.

\begin{table}[t]
\normalsize
\centering
\begin{tabular}{c|c}
    \hline
    {Method}	&	{Seconds per token }\\
    \hline
    Adaptive-wait-$k$ &	0.1057 s	\\
    PED	&	0.0358 s	\\
    PED w/o PE	&	0.0175 s\\
    Wait-$k$	&	0.0146 s	\\
    \hline
\end{tabular}
\caption{The comparison of average time to generate a target token in different methods.
}
\label{time}
\end{table}

\subsection{Translation Efficiency}
In order to compare the translation efficiency between our method and the previous methods, we measure it by using the average time of generating each token. The results in Table \ref{time} are tested on GeForce GTX TITAN-X.  It can be seen that the translation speed of our methods is less than wait-$k$ policy, but about three times faster than Adaptive-wait-$k$ policy. Besides, the translation speed of PED is about twice as slow as `PED w/o PE', which is roughly in line with our expectation for our Post-Evaluation policy.

\section{Related Work}
\textbf{SiMT} policy can be divided into fixed and adaptive policy according to whether the translation path is dynamically decided based on context. For fixed policy, the number of READ actions between adjacent WRITE actions always keeps constant. \citet{DBLP:conf/naacl/DalviDSV18} proposed STATIC-RW, and \citet{DBLP:conf/acl/MaHXZLZZHLLWW19} proposed wait-$k$ policy, which reads and writes a token alternately after reading $k$ tokens. \citet{multiPath} proposed multi-path training method to make a unified model perform multiple wait-$k$ policies and get the performance comparable to the wait-$k$ policy under all latency. \citet{DBLP:conf/aaai/ZhangFL21} proposed future-guided training to help SiMT model invisibly embed future information via knowledge distillation. \citet{zhang-feng-2021-icts} proposed a char-level wait-k policy to improve the robustness of SiMT. \citet{zhang-feng-2021-universal} proposed MoE wait-k policy, which treats the attention heads as a set of wait-$k$ experts, thereby achieving state-of-the-art performance among the fixed policies.

For adaptive policy, \citet{DBLP:conf/emnlp/ZhengZMH19} trained the agent with oracle actions generated by full-sentence neural machine translation model. \citet{arivazhagan-etal-2019-monotonic} proposed MILk to decide the READ and WRITE actions by introducing a Bernoulli variable. \citet{DBLP:conf/iclr/MaPCPG20} proposed MMA, which implemented MILk on Transformer. \citet{DBLP:conf/acl/ZhengLZMLH20} implemented the adaptive policy through a composition of several fixed policies. \citet{miao-etal-2021-generative} proposed a generative framework to perform the adaptive policy for SiMT. \citet{DBLP:conf/acl/ZhangF22}  introduced duality constraints to direct the learning of translation paths during training. Instead of predicting the READ and WRITE actions, \citet{zhang-feng-2022-gaussian} implemented the adaptive policy by predicting the aligned source positions of each target token.

Our method focuses on the accuracy of READ and WRITE actions during inference. Our PE policy can evaluate the rationality of actions by utilizing the increment of source content before taking them, which reduces the probability of incorrect actions. Besides, our method achieves good performance under all latency with a unified model.

\textbf{Capsule Networks} \citep{capNet} and its assignment policies \citep{DBLP:conf/nips/Sabour, DBLP:conf/iclr/HintonSF18} initially attempted to solve the problem of parts-to-wholes in computer vision. \citet{DBLP:conf/aaai/DouTWWSZ19} first employed capsule network into NMT (i.e, neural machine translation) model for layer representation aggregation. \citet{zheng-etal-2019-dynamic} proposed a novel assignment policy GDR to model past and future source content to assist translation. \citet{wang-2019-towards} proposed a novel capsule network for linear time NMT. 

Our PED method introduces Capsule Networks with GDR into SiMT model and recognizes the translation degree of source tokens under the restriction of partial source information. Furthermore, we evaluate the rationality of the actions by measuring the changes in translation degree, to implement the adaptive policy.

\section{Conclusion}
In this paper, we propose a new method of performing the adaptive policy by integrating post-evaluation into the fixed policy to evaluate the rationality of the actions. Besides, disturbed-path training is proposed to enhance the robustness of the model to recognize the translation degree on different translation paths. Experiments show that our method outperforms the strong baselines under all latency and can recognize the translation degree on different paths accurately. Furthermore, PE policy can enhance the sufficiency and necessity of translation paths to achieve better performance.

\section*{Limitations}
We think our methods mainly have two limitations. On the one hand, although our method can recognize the translation degree of each source token, it still has some deviations. On the other hand, although the inference speed of our method is slightly slower than the wait-$k$ policy, it is still faster than the Adaptive-wait-$k$ policy, which is enough to meet the needs of the application.

\section*{Acknowledgements}
We thank all the anonymous reviewers for their insightful and valuable comments.

\bibliography{anthology,custom}
\bibliographystyle{acl_natbib}

\appendix

\begin{table*}[t]
\centering
\begin{tabular}{l c c c}
\toprule
\textbf{Hyperparameter} & \textbf{IWSLT15 En$\rightarrow $Vi} & \textbf{IWSLT14 En$\rightarrow $De} & \textbf{WMT15 De$\rightarrow $En}\\ \hline
encoder layers          & 6         & 6             &6                \\
encoder attention heads & 4         & 4             &8                \\
encoder embed dim       & 512       & 512           &512                \\
encoder ffn embed dim   & 1024      & 1024          &2048                \\
decoder layers          & 6         & 6             &6                \\
decoder attention heads & 4         & 4             &8                \\
decoder embed dim       & 512       & 512           &512                \\
decoder ffn embed dim   & 1024      & 1024          &2048                \\
dropout                 & 0.3       & 0.3           &0.3                \\
optimizer               & adam      & adam          &adam                \\
adam-$\beta$          & (0.9, 0.98) & (0.9, 0.98)   & (0.9, 0.98)     \\
clip-norm               & 0         & 0             &0                \\
lr                      & 5e-4      & 5e-4          & 5e-4                \\
lr scheduler        & inverse sqrt  & inverse sqrt  & inverse sqrt     \\
warmup-updates          & 4000      & 4000          & 4000            \\
warmup-init-lr          & 1e-7      & 1e-7          & 1e-7                \\
weight decay            & 0.0001    & 0.0001        & 0.0001        \\
label-smoothing         & 0.1       & 0.1           & 0.1                \\
max tokens              & 16000     &  8192$\times$4 & 2048$\times$4$\times$4 \\
\bottomrule
\end{tabular}
\caption{Hyperparameters of our experiments.}
\label{Hyper}
\end{table*}

\section{Hyperparameters}
All systems in our experiments use the same hyperparameters, as shown in Table \ref{Hyper}.

\section{Numerical Results}
Table \ref{envi}, \ref{deen}, \ref{ende} respectively report the numerical results on IWSLT15 En$\rightarrow $Vi, IWSLT14 En$\rightarrow $De and WMT15 De$\rightarrow$En measured by AL and BLEU.

\begin{table}[]
\centering
\begin{tabular}{p{2cm}<{\centering} p{2cm}<{\centering} p{2cm}<{\centering}} 
\toprule[1.5pt]
\multicolumn{3}{c}{\textbf{IWSLT15 En$\rightarrow $Vi}}                \\
\midrule[1pt]
\multicolumn{3}{c}{\textit{\textbf{Offline}}}     \\\hline
          & AL  & BLEU   \\
          & 22.41    &28.8   \\
\midrule[1pt]
\multicolumn{3}{c}{\textit{\textbf{Wait-$k$}}}     \\
\hline
    $k$      & AL  & BLEU   \\
    1      &3.03      &25.28  \\
    3      &4.64      &27.53  \\
    5      &6.46      &28.27  \\
    7      &8.11      &28.45  \\
    9      &9.80      &28.53  \\
\midrule[1pt]
\multicolumn{3}{c}{\textit{\textbf{Multi-path}}}     \\
\hline
    $k$      & AL  & BLEU   \\
    1      &3.16     &25.82   \\
    3      &4.69     &27.99  \\
    5      &6.42     &28.33  \\
    7      &8.17     &28.39  \\
    9      &9.82     &28.36 \\
\midrule[1pt]

\multicolumn{3}{c}{\textit{\textbf{Adaptive-wait-$k$}}}     \\
\hline
    ($\rho_1$, $\rho_{10}$)    & AL  & BLEU   \\
    (0.2, 0.0)      &3.12       &26.05   \\
    (0.4, 0.0)      &4.38       &27.72  \\
    (0.6, 0.0)      &6.28       &28.45  \\
    (1.0, 0.0)      &7.96       &28.47  \\
    (1.0, 0.4)      &9.80       &28.41  \\
\midrule[1pt]

\multicolumn{3}{c}{\textit{\textbf{PED}}}     \\
\hline
    $k$      & AL  & BLEU   \\
    1      &3.16      &26.78   \\
    3      &4.74      &28.69  \\
    5      &6.46      &28.74  \\
    7      &8.18      &28.82  \\
    9      &9.80      &28.77 \\
\midrule[1pt]
\end{tabular}
\caption{Numerical results of IWSLT15 En$\rightarrow$Vi.}
\label{envi}
\end{table}

\begin{table}[]
\centering
\begin{tabular}{p{2cm}<{\centering} p{2cm}<{\centering} p{2cm}<{\centering}} 
\toprule[1.5pt]
\multicolumn{3}{c}{\textbf{IWSLT14 En$\rightarrow $De}}                \\
\midrule[1pt]
\multicolumn{3}{c}{\textit{\textbf{Offline}}}     \\\hline
          & AL  & BLEU   \\
          & 23.25    &27.18   \\
\midrule[1pt]
\multicolumn{3}{c}{\textit{\textbf{Wait-$k$}}}     \\
\hline
    $k$      & AL  & BLEU   \\
    1      &2.03      &18.54  \\
    3      &3.31      &22.30  \\
    5      &5.17      &25.45  \\
    7      &6.83      &26.01  \\
    9      &8.52      &25.64  \\
\midrule[1pt]
\multicolumn{3}{c}{\textit{\textbf{Multi-path}}}     \\
\hline
    $k$      & AL  & BLEU   \\
    3      &3.22     &23.50  \\
    5      &5.01     &25.84  \\
    7      &6.84     &26.65  \\
    9      &8.64     &26.83 \\
\midrule[1pt]

\multicolumn{3}{c}{\textit{\textbf{Adaptive-wait-$k$}}}     \\
\hline
    ($\rho_1$, $\rho_{10}$)    & AL  & BLEU   \\
    (1.0, 0.3)      &2.34       &24.08  \\
    (1.0, 0.4)      &3.79       &24.63  \\
    (1.0, 0.6)      &6.34       &25.74  \\
    (1.0, 0.7)      &7.07       &25.88  \\
    (1.0, 0.8)      &8.10       &26.07  \\
\midrule[1pt]

\multicolumn{3}{c}{\textit{\textbf{PED}}}     \\
\hline
    $k$      & AL  & BLEU   \\
    3      &3.05      &24.14  \\
    5      &5.03      &26.16  \\
    7      &6.91      &26.81  \\
    9      &8.71      &27.12 \\
\midrule[1pt]
\end{tabular}
\caption{Numerical results of IWSLT14 En$\rightarrow$De.}
\label{ende}
\end{table}

\begin{table}[]
\centering
\begin{tabular}{p{2cm}<{\centering} p{2cm}<{\centering} p{2cm}<{\centering}} 
\toprule[1.5pt]
\multicolumn{3}{c}{\textbf{WMT15 De$\rightarrow $En}}                \\
\midrule[1pt]
\multicolumn{3}{c}{\textit{\textbf{Offline}}}     \\\hline
          & AL  & BLEU   \\
          & 27.45    &30.62   \\
\midrule[1pt]
\multicolumn{3}{c}{\textit{\textbf{Wait-$k$}}}     \\
\hline
    $k$      & AL  & BLEU   \\
    1      &-0.01     &17.88  \\
    3      &1.66      &23.23  \\
    5      &4.12      &26.88  \\
    7      &6.01      &28.35  \\
    9      &7.84      &28.97  \\
\midrule[1pt]
\multicolumn{3}{c}{\textit{\textbf{Multi-path}}}     \\
\hline
    $k$      & AL  & BLEU   \\
    1      &0.64      &19.90   \\
    3      &2.20       &24.06  \\
    5      &4.10       &26.87  \\
    7      &6.08      &28.46  \\
    9      &8.00         &29.42 \\
\midrule[1pt]

\multicolumn{3}{c}{\textit{\textbf{Adaptive-wait-$k$}}}     \\
\hline
    ($\rho_1$, $\rho_{10}$)    & AL  & BLEU   \\
    (0.2, 0.0)      &0.50       &20.37   \\
    (0.4, 0.0)      &1.39       &22.81  \\
    (0.6, 0.0)      &2.52       &25.28  \\
    (0.8, 0.0)      &4.39       &27.63  \\
    (1.0, 0.0)      &5.38       &28.15  \\
    (1.0, 0.4)      &7.32       &28.78 \\
\midrule[1pt]

\multicolumn{3}{c}{\textit{\textbf{PED}}}     \\
\hline
    $k$      & AL  & BLEU   \\
    1      &- 0.21    &22.08   \\
    3      &1.62      &24.57  \\
    5      &3.39      &27.51  \\
    7      &5.67      &29.16  \\
    9      &7.63      &30.28 \\
\midrule[1pt]
\end{tabular}
\caption{Numerical results of WMT15 De$\rightarrow$En.}
\label{deen}
\end{table}

\end{document}